\documentclass{article} % review

\usepackage{authblk}

\usepackage{amsmath}
\usepackage{amsfonts}
\usepackage{amssymb}
\usepackage{bm}
\usepackage{subcaption}
\usepackage{color}
\usepackage{multirow}
\usepackage{adjustbox}
\usepackage{booktabs}
\usepackage{placeins}

\renewcommand{\vec}{\text{vec}} 
\newcommand{\R}{\mathbb{R}} 
\newcommand{\C}{\mathbb{C}} 
\newcommand{\diag}{\text{diag}} 
\newcommand{\norm}[1]{\left\lVert#1\right\rVert}

\newcommand{\range}[2]{{#1},\ldots, {#2}}
\newcommand{\hinf}{\mathcal{H}_{\infty}}
\newcommand{\regpar}{\gamma}
\newcommand{\regfun}{\mathcal{R}}
\newcommand{\eig}{\mathtt{eig}}

\newcommand{\nx}{{n_x}} 
\newcommand{\ny}{{n_y}} 
\newcommand{\nin}{{n_u}} 
\newcommand{\dmodel}{{d_{\rm model}}}
\newcommand{\nlayers}{{n_{\rm layers}}}

\usepackage{layouts}

\usepackage{url}

\newtheorem{remark}{Remark}

\title{Model order reduction of deep structured state-space models: A system-theoretic approach} 
\author{Marco Forgione}
\author{Manas Mejari}
\author{Dario Piga}
\affil{IDSIA Dalle Molle Institute for Artificial Intelligence USI-SUPSI, Via la Santa 1, CH-6962 Lugano-Viganello, Switzerland.}

\begin{document}
\maketitle

\begin{abstract}
With a specific emphasis on control design objectives, achieving accurate system modeling with limited complexity is crucial in parametric system identification.  
The recently introduced deep structured state-space models (SSM), which feature linear dynamical blocks as key constituent components, offer high predictive performance. However, the learned representations often suffer from excessively large model orders, which render them unsuitable for control design purposes.
The current paper addresses this challenge  by means of system-theoretic model order reduction techniques that target the linear dynamical blocks of SSMs. We introduce two regularization  terms which can be incorporated into the training loss for improved model order reduction. In particular, we consider modal $\ell_1$ and Hankel nuclear norm  regularization to promote sparsity, allowing one to retain only the relevant states without sacrificing accuracy. The presented regularizers lead to advantages in terms of parsimonious representations and faster inference resulting from  the reduced order models.
The effectiveness of the proposed methodology is demonstrated using real-world ground vibration data from an  aircraft.
\end{abstract}

\section{Introduction}
In recent years, deep structured state-space models (SSM) have emerged as  powerful and flexible architectures to tackle machine-learning tasks over sequential data such as time series classification, regression, and forecasting~\cite{Gu22, Gu21,  Orvieto23, Smith22}. Notably, they exhibit state-of-the-art performance in problems defined over very long sequences, where Transformers struggle due to their computational complexity that grows quadratically with the sequence length~\cite{Tay2020}. 

Essentially, SSMs consist in the sequential connection of linear \emph{dynamical} blocks interleaved with static non-linearities and normalization units, organized in identical repeated layers with skip connections (see Fig.~\ref{fig:deep_lru_architecture}). In this sense, they are closely related to the block-oriented modeling framework~\cite{ schoukens2017identification} traditionally employed by the system identification community, and  made compatible for training  in a deep-learning  setting thanks to the \emph{dynoNet} architecture proposed by some of the authors in~\cite{forgione21}. 

Several mathematical and software implementation solutions have been devised to make learning of SSM architectures--in particular of their key constituent linear dynamical block--fast, well-posed, and efficient. For instance, S4 \cite{Gu21} adopts a continuous-time parameterization, an initialization strategy based on continuous-time memorization theory, and a convolution-based implementation in frequency domain based on fast Fourier transform. Conversely, the deep Linear Recurrent Unit architecture~\cite{Orvieto23} adopts a discrete-time parameterization with a diagonal state-transition matrix, an initialization strategy that constrains the eigenvalues of the system in a region of stability, and an efficient implementation in time domain exploiting the parallel scan algorithm~\cite{blelloch1990prefix}.

From the Systems and Control perspective, \emph{parsimonious} representations are often sought, \emph{i.e.}, it is desired to obtain a model that describes system's behaviour with as few parameters and states as possible, to simplify downstream tasks such as controller synthesis, state estimation, etc. 
% Furthermore, while complex models such as deep SSMs 
% may be able to capture intricate dynamics accurately,  there is a risk of over-fitting and high variance in the estimated model parameters, notably for small training datasets. 
% It is therefore paramount to limit the model complexity to improve the model's generalization capabilities.

{The inadequacy of high-dimensional models has driven growing interest in Model Order Reduction (MOR) techniques. In particular, several contributions focus on reducing the number of states of linear dynamical systems, employing methods that can be broadly categorized into SVD-based and Krylov approximation-based techniques~\cite{ath05}.  The former rely on the concept of the \emph{Hankel singular values}, which characterize the complexity of the reduced-order model and provide an error bound of the approximation~\cite{glover84}. These methods include balanced truncation~\cite{alsaggaf88, moore81}, singular perturbation approximation~\cite{liu89spa} and  Hankel norm approximation~\cite{pillonetto2016}. On the other hand, Krylov-based approximation methods are iterative in nature. They are based on \emph{moment matching} of the impulse response rather than computation of singular values, see the recent survey paper~\cite{SCARCIOTTI2024moment} for an overview of these approaches. 
}

In this paper, we demonstrate the effective adaptation of these MOR techniques, initially designed for linear dynamical systems, to the task of learning simplified deep SSMs architectures while maintaining their predictive capabilities. In principle, an SSM could  first be learned using standard machine-learning algorithms, and then each of the constituent linear dynamical blocks could be reduced employing one of the MOR techniques mentioned above.
However, we show that the effectiveness of MOR is significantly increased when the low-order modeling objective is already integrated in the training procedure, by means of a \emph{regularization term} in the loss function which promotes parsimonious representations.
In particular, we adopt modal $\ell_1$ and Hankel nuclear norm regularization approaches that penalize the magnitude of the linear units' eigenvalues and Hankel singular values, respectively.
We illustrate our methodology on a well-known system identification benchmark~\cite{noel2017f} where the aim is to model the oscillations of an aircraft subject to a ground vibration test. We show that specific combinations of regularizers applied during training, along with  MOR techniques applied after training, yield the best results.
%\footnote

{All our codes are available in the GitHub repository \url{https://github.com/forgi86/lru-reduction}, allowing full reproducibility of the reported results.}

\section{Problem Setting}\label{sec:problem}

We consider a training dataset consisting of a sequence of input-output samples $\mathcal{D}= \{u_k, y_k \}_{k=1}^{N}$, generated from an unknown dynamical system $\mathcal{S}$, where $u_k \in \R^{\nin}$ is the input and $y_k \in \R^{\ny}$ is the measured output at time step $k$.
The problem considered in this work is to learn 
 a parametric simulation model $\mathcal{M}(\theta)$ with parameters $\theta \in \Theta$, mapping an input sequence $u_{1:k}$ to the (estimated) output sequence $\hat{y}_{1:k}$, which fits the training dataset $\mathcal{D}$. In particular, the focus is  to identify a \emph{parsimonious} model with as few states (in turn, parameters) as possible  via regularization and model order reduction techniques.

The parameters  $\theta$  characterising the model $\mathcal{M}(\theta)$ are estimated according to the criterion:
\begin{align}
\hat{\theta} =  \arg\min_{\theta \in \Theta}  & \frac{1}{N}\sum_{k=1}^N \mathcal{L}\left(y_k, \hat{y}_k(\theta) \right) +  \regpar \regfun(\theta),    \label{eqn:lossE}
\end{align}
where $\hat{y}_k(\theta)$ represents the model's output at time $k$, and $\mathcal{L}(\cdot)$ denotes the chosen fitting loss function. The term $\regfun(\theta)$ is a regularization cost which aims at enforcing sparsity and reducing the complexity of the model, ultimately facilitating the subsequent MOR step. Different choices for the regularization loss $\regfun(\theta)$ will be introduced in the Section~\ref{sec:reg}.

In this work, $\mathcal{M}(\theta)$ is a SSM architecture recently introduced in~\cite{Orvieto23} and known as deep Linear Recurrent Unit. In the next section, we describe in details the building blocks and parameterizations of this architecture.

\section{Deep Structured State-Space Model}\label{sec:lru}
\begin{figure}
\centering
\includegraphics[width=.5\linewidth]{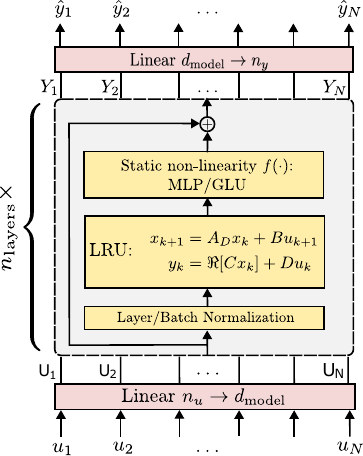}
\caption{The Deep Linear Recurrent Unit architecture.}
\label{fig:deep_lru_architecture}
\end{figure}
 The deep Linear Recurrent Unit architecture  is visualized in Fig.~\ref{fig:deep_lru_architecture}. Its core (shown in gray) is a stack of $\nlayers$ Linear Recurrent Units (LRU), which are linear dynamical systems, interleaved with static non-linearities, (\emph{e.g.}, Multi Layer Perceptrons (MLP)  or Gated Linear Units (GLU)~\cite{dauphin2017language}) and normalization units (typically Layer or Batch Normalization~{\cite{Wu2018}}), with \emph{skip connections} included in each repeated layer.  In particular, the $l$-th layer of the network with inputs $u^{l}_k $ and output $s^{l}_k $ is defined by:
 \begin{equation}
  \mathcal{M}^{l}:
    \begin{cases}
      \tilde{u}^{l}_k = {\rm{Norm}}({u}^{l}_k),\\
      y^{l}_k = \mathrm{LRU}(\tilde{u}^{l}_{1:k};\theta^l),  \\
      s^{l}_k = u^{l}_k + f(y^{l}_k), \\
      \end{cases}       
\end{equation}
where $\rm{Norm}(\cdot)$ is the normalization unit; the LRU is a linear time-invariant (LTI) multi-input multi-output (MIMO) dynamical block whose exact structure is described in the next sub-section; and $f(\cdot):\R^\dmodel \rightarrow \R^\dmodel$ is the static non-linearity, applied in an element-wise fashion to all samples of the LRU output sequence. 
The first and last transformations in the architecture (pink blocks) are static linear projections mapping the input samples $u_k \in \R^{\nin}$ to $U_k \in \R^\dmodel$, and $Y_k \in \R^\dmodel$ to predicted outputs $\hat y_k \in \R^{\ny}$, respectively.

% The architecture above has been introduced in~\cite{Gu21} and then adopted without substantial modifications in the deep SSM literature, except for the internal structure of the  linear dynamical block.
We  remark that deep LRU shares a close proximity to the \emph{dynoNet} architecture
proposed in~\cite{forgione21}. The main difference is that the LRU is a state-space representation of an LTI system, while \emph{dynoNet} employs input-output transfer function descriptions. 
The architecture is also related to (decoder-only) Transformers~\cite{Radford2019}, with information shared across time steps with an LTI system instead of a causal attention layer.

In the following subsection, we summarize the details of the LRU block. We omit the layer index $l$ when referencing parameters and signals to simplify the notation. Furthermore, %with a slight abuse of notation,
we redefine $u_k/y_k$ as the input/output samples of the LRU to match the standard notation of linear system theory.

\subsection*{Linear Recurrent Unit}
The LRU is a linear, discrete-time, MIMO dynamical system described in state-space form as:
\begin{subequations}\label{eq:lru_ss}
\begin{align}
    x_k &= A_D \ x_{k-1} + B u_k, \\
    y_k &= \Re[{C x_k}] + D u_k,
\end{align}
\end{subequations}
where $\Re[\cdot]$ denotes the real part of its argument, $A_D \in \C^{\nx \times \nx}$, $B \in \C^{\nx \times \nin}$ and $C \in \C^{\ny \times \nx}$ are  complex-valued matrices, and $D \in \R^{\ny \times \nin}$ is a real-valued matrix. 
The matrix $A_D$ has a diagonal structure:
\begin{equation}\label{eq:diagA}
A_D = \diag(\lambda_1, \ldots, \lambda_{\nx}),
\end{equation}
where $\lambda_j, \ j=\range{1}{\nx}$ are the complex eigenvalues, or {modes}, of the system, which is thus represented in a \emph{modal} form.
In order to guarantee  
asymptotic stability of the system, \emph{i.e}, to enforce  
$|\lambda_j| <1, \  j = \range{1}{\nx}$, each  eigenvalue $\lambda_j \in \mathbb{C}$ is, in turn, parameterized as
$
    \lambda_j = \exp(-\exp(\nu_j)+i \exp(\phi_j)),
$
where $\nu_j > 0$. Note that since $\exp(\nu_j)>0$ for $\nu_j > 0$, this ensures  $|\lambda_j|= \exp(-\exp(\nu_j))<1$. 

The input matrix $B$ is  parameterized as  
$B = \diag(\gamma_{1},\ldots,\gamma_{\nx})\tilde{B}$, where $\gamma_{j}= \sqrt{1-|\lambda_j|^2}, \ j = \range{1}{\nx}$, is a normalization factor introduced to obtain state signals with the same power as that of the input signal. 
Overall, the learnable parameters of a single LRU block are $\theta :=\{ \{\nu_j,\phi_j\}_{j=1}^{\nx}, \tilde{B}, C, D\}$.  
\begin{remark}
    System \eqref{eq:lru_ss} has an equivalent complex-conjugate representation:
\begin{subequations}\label{eq:lru_ss_cc}
\begin{align}
    \tilde x_k &= \begin{bmatrix}A_D & 0 \\ 0 & A_D^* \end{bmatrix} \ \tilde x_{k-1} + \begin{bmatrix}B\\B^* \end{bmatrix} u_k, \\
    y_k &= \frac{1}{2}\begin{bmatrix}C\;\; C^*\end{bmatrix}\tilde x_k + D u_k,
\end{align}
with $\tilde{x}_k \in \C^{2\nx}$, which in turn may be transformed in a real Jordan form, with a block-diagonal state-transition matrix containing $\nx$ 2x2 blocks, see e.g. Appendix E.3 of~\cite{Orvieto23}. The complex-valued, diagonal representation~\eqref{eq:lru_ss} is preferred for its implementation simplicity and halved state dimension.
\end{subequations}
\end{remark}

\section{Model Order Reduction and Regularization}\label{sec:MOR_reg}
In this section, we provide a brief review of the MOR techniques used in this paper. Next, we introduce regularization techniques  aimed at promoting the learning of parsimonious LRU representations in terms of state complexity.%, which can then be effectively simplified offline through MOR techniques.

\subsection{Reduction by truncation and singular perturbation}\label{sec:MOR}
Order reduction techniques based on truncation decrease the dimensionality of a dynamical system by eliminating states that, for various reasons, are considered less important~\cite{green12}. Consider a state-space model $G$ with realization:
\begin{subequations}  
\label{eq:ss_partitioned}
\begin{align}\label{eq:ss_partitioned_a}
    \begin{bmatrix}
    x_{1,k}\\
    x_{2,k}
    \end{bmatrix}
    &=
    \begin{bmatrix}A_{11} & A_{12} \\ A_{21} & A_{22} \end{bmatrix}     
    \begin{bmatrix}
    x_{1,k-1}\\
    x_{2,k-1}
    \end{bmatrix} + \begin{bmatrix}B_1\\B_2 \end{bmatrix} u_k, \\
    y_k &= \begin{bmatrix}C_1\;\; C_2\end{bmatrix} \begin{bmatrix}x_{1,k}\\ x_{2,k} \end{bmatrix} + D u_k,
\end{align}
\end{subequations}
where the partitioning corresponds to important states to be kept $x_1 \in \R^r$ and unimportant states to be removed $x_2 \in \R^{\nx - r}$, respectively. The state-space \emph{truncation} method approximates~\eqref{eq:ss_partitioned} with a reduced-order system $G_r$ having state-space matrices:
\begin{equation}
\label{eq:balanced_truncation}
A_r = A_{11}, B_r = B_1, C_r = C_1, D_r = D.
\end{equation}
Note that state-space truncation does not preserve the steady-state behavior of $G$
%of the original system~\eqref{eq:ss_partitioned} 
and, in general, it may alter its low-frequency response significantly. 
Alternatively, the removed states $x_2$ may be set to equilibrium by solving for $x_{2,k}= x_{2,k-1}$ in~\eqref{eq:ss_partitioned_a}. This results in the so-called \emph{singular perturbation} approximation, where the reduced-order system $G_r$ is defined by the following state-space matrices:
\begin{subequations}   
\label{eq:singular_perturbation}
\begin{align}
    A_r &= A_{11} + A_{12}(I - A_{22})^{-1}A_{21}\\
    B_r &= B_1 + A_{12} (I - A_{22})^{-1} B_2 \\
    C_r &= C_1 + C_2 (I - A_{22})^{-1} A_{21} \\
    D_r &= C_2 (I - A_{22})^{-1} B_2 + D.
\end{align}
\end{subequations}
Singular perturbation preserves the steady-state behavior of~\eqref{eq:ss_partitioned}, and generally provides a better match in the lower frequency range with respect to plain state-space truncation.

In the control literature, state-space truncation and singular perturbation approximations are typically applied to systems described either in modal or in \emph{balanced} realization form~\cite{green12}. The resulting MOR techniques will be denoted here as modal truncation (MT), modal singular perturbation (MSP), balanced truncation (BT), and balanced singular perturbation (BSP).
In the modal methods, the states are sorted according to non-increasing magnitude of the corresponding eigenvalues, so that the fastest dynamics can be discarded. This choice is often motivated by physical arguments, when the fast dynamics are associated to uninteresting second-order effects (e.g., electrical dynamics in mechanical systems). In balanced methods, the states are sorted for non-increasing value of the corresponding Hankel singular values. This choice is supported by the following approximation bound, which holds both for BT and BSP:
\begin{equation}
\label{eq:hinf_error_bound}
\norm{G-G_r}_{\infty} \leq 2\sum \limits_{j=r+1}^{\nx}\sigma_j,
\end{equation}
where $\norm{\cdot}_{\infty}$ denotes the $\hinf$ norm and $\sigma_j, \ j=\range{r+1}{\nx}$ are the Hankel singular values corresponding to the eliminated states, see~\cite[Lemma $3.7$]{katayama05}.

For the LRU, MT/MSP are directly applicable, being the block already represented in a modal form. Conversely, BT/BSP require a three-step procedure where (i) system~\eqref{eq:lru_ss} is first transformed to a (non-diagonal) balanced form, then (ii) reduced according to either~\eqref{eq:balanced_truncation} or~\eqref{eq:singular_perturbation}, and finally (iii) re-diagonalized with a state transformation obtained from the eigenvalue decomposition of the state matrix of the system obtained at step (ii) to fit the LRU famework.

\subsection{Regularized Linear Recurrent Units} \label{sec:reg}
In this section, we introduce two {regularization} approaches that promote learning of LRUs  with a reduced  state complexity. 
These methods leverage system-theoretic MOR methods described in the previous sub-section and  exploit the diagonal structure of the LRU's state-transition matrix $A_{D}$. 
\subsubsection{Modal $\ell_1$-regularization}

%In physical systems, 
As mentioned in Section~\ref{sec:MOR}, high-frequency modes often correspond to secondary phenomena that may be eliminated without compromising the salient aspects of the modeled dynamics.
%behaviours % resulting
%in such modes only 
%$play a secondary role in determining the system dynamics.
%In discrete-time systems, modes with a fast decay rate are associated to eigenvalues with modulus close to zero.
%as they represent unrealistic or uninteresting effects.
%In several physical systems, it is often the case that high-frequency modes corresponding to fast dynamics may be ignored as they represent uninteresting effects or such modes do not play a significant role in the overall input-output behaviour. % of the model.  
For discrete-time LTI systems, fast dynamics are associated to eigenvalues whose modulus $|\lambda_j|$ is small. 
%This rationale can be incorporated into training 
An $\ell_1$-regularization is therefore introduced to push some of the modes towards zero during training:
%them from the system at a second
%Thus, their effect can be eliminated 
%by adding a LASSO regularization term  based on the $\ell_1$-norm of the eigenvalues of $A_d$ as follows: 
%In this sense, an effective regularized which
\begin{equation}
    \label{eq:lasso}
    \regfun(\theta) = \sum_{l=1}^{\nlayers} \sum_{j=1}^{\nx} |\lambda_j^{l}|.
\end{equation}
Indeed, $\ell_1$-regularization is known to promote sparsity in the solution~\cite{tibshirani1996lasso}. The states corresponding to eigenvalues that are closer to zero 
will then be eliminated with a MOR method  at a second stage after training.  Note that modal $\ell_1$-regularization  of the LRU is computationally cheap, as the eigenvalues  are directly available on the diagonal of the
state-transition matrix $A_D$. 

\subsubsection{Hankel nuclear norm regularization}

 %The Markov parameters or the impulse response coefficients $g_k$ of a discrete-time LTI system are given by $g_k = CA_{d}^{k-1}B, \ \forall k \in \range{1}{\infty}$. 
%The Hankel operator $H(g)$ associated with these impulse response coefficients $g_k$  is given as follows,
% \begin{equation}\label{eq:Hankel_matrix}
%     H(g) =   \begin{bmatrix}
% g_1 &g_2  & g_3 & \cdots \\
%  g_2& g_3 & g_4 & \cdots\\
%  g_3& g_4 &  g_5& \cdots\\ 
%  \vdots & \vdots &\vdots &\ddots
% \end{bmatrix}
% \end{equation}

It is known that the  McMillan degree (minimum realization order)
of a discrete-time LTI system  coincides with the rank 
of its associated (infinte-dimensional) Hankel operator $H$~\cite{katayama05}. The $(i,j)$-th block of $H$ is  defined as $H_{ij} = g_{i+j-1}$, where $g_k= CA_{D}^{k-1}B$ is the impulse
response coefficient at time step $k$. Minimizing the rank of the Hankel operator thus aligns with the objective of obtaining a low-order representation. However, the rank minimization problem is hard to solve and  the \emph{nuclear norm} of the Hankel operator $\norm{H(g)}_{*} := \sum_{j} \sigma_j$, defined as the sum of its singular values $\sigma_j$, is often  used as a \emph{convex surrogate} of the rank~\cite{fazel2001, pillonetto2016}.

%Note that the $\hinf$ error bounds defines the error between the output

Following this rationale, we employ the Hankel nuclear norm of the LRUs as a regularization term in training:
\begin{equation}
    \label{eq:hankel_reg}
    \regfun(\theta) = \sum_{l=1}^{\nlayers} \sum_{j=1}^{\nx} \sigma_j^l,
\end{equation}
where $\sigma_j^l$ denotes the $j$-th singular value of the Hankel operator of the LRU in the $l$-th layer. Note that, as $\sigma^{l}_j \geq 0,\  j = \range{1}{\nx}$, the term $\sum_{j=1}^{\nx} \sigma_j^l$ in \eqref{eq:hankel_reg} is the $\ell_1$-norm of the Hankel singular values, thus, promoting sparsity.  

It can be proved that the $j$-th singular value of the Hankel operator is given by $\sigma_j(H) = \sqrt{\eig_j(PQ)}$, where $P$ and $Q$ are the controllability and observability Grammians of the LTI model~\cite{katayama05}. 
In appendix~\ref{sec:app}, we show how the Grammians $P$ and $Q$, and in turn the Hankel singular values can be computed efficiently for systems in modal form.
\begin{remark}[$\ell_2$-regularization]\label{rem:L2_reg}
If the $\ell_2$-norm of the Hankel singular values is considered in  \eqref{eq:hankel_reg} instead of the $\ell_1$-norm, the computational burden  during  training can be further reduced exploiting the identity $\sum_{j=1}^{\nx} \sigma^{2}_j = \sum_{j=1}^{\nx} \eig_j(PQ)= \mathrm{trace}(PQ)$. Thus,  differentiation  of the eigenvalues of $PQ$ is not required. Nonetheless, it is known that $\ell_2$-norm regularization does not enforce  sparsity in the solution, contrary to the $\ell_1$ case. 
\end{remark}

\begin{remark}[$\hinf$-error bound]
The use of the regularizer~\eqref{eq:hankel_reg} is further motivated by the $\hinf$ error bound \eqref{eq:hinf_error_bound}. This suggests to combine Hankel-norm regularization during training with MOR based on either BT or BSP.
\end{remark}

\section{Case Study}\label{sec:examples}

We test the methodologies of the paper on the ground vibration dataset of an F-16 aircraft introduced in~\cite{noel2017f}.

The input $u \in \R$~(N) is the force  generated by a shaker mounted on the aircraft's right wing, while the outputs $y \in \R^3$~($m/s^2$) are the accelerations measured at three test locations on the aircraft. Input and output samples are collected at a constant frequency of $400$~Hz.
We train deep LRU models with structure as shown in Fig.~\ref{fig:deep_lru_architecture} characterized by: Layer Normalization; MLP non-linearity; $\dmodel = 50$; $\nx=100$; and $\nlayers=6$. The MLP has one hidden layer with 
%$4\times \dmodel = 400$
400
hidden units and Gaussian Error Linear Unit (GELU) non-linearities.%\footnote{This MLP structure with ``branching factor'' 4 and GELU non-linearity is a common default choice in Transformer architectures, see e.g. ~\cite{Radford2019}.} 

Training is repeated three times with identical settings except for the regularization strategy, which is set to: (i) no regularization, (ii) modal $\ell_1$-regularization, and (iii) Hankel nuclear norm regularization. For both (ii) and (iii), the regularization strength  is set to $\gamma=10^{-2}$. 
%\footnote{Exploration of different values of $\lambda$ will be the subject of future investigations.}
We train the models on all the input/output sequences suggested for training in~\cite{noel2017f} except the one denoted as ``Special Odds''. To take advantage of parallel computations on more sequences, we split the datasets in (partially overlapping) sub-sequences of length $N=5000$ samples each and compute the training loss~\eqref{eqn:lossE} over batches of $64$ sub-sequences simultaneously. In the definition of the loss, the first 200 samples of each sub-sequence are discarded to cope with the state initialization issue, according to the ZERO initialization scheme described in~\cite{forgione2022learning}. We minimize the mean squared error training loss over $10$ epochs of AdamW with constant learning rate $10^{-4}$, where at each epoch all the 688820 sub-sequences of length $N$ in the training data are processed. 

We report the \emph{fit} index~\cite{schoukens2019nonlinear}, the Root Mean Squared Error (RMSE), and the Normalized Root Mean Squared Error (NRMSE) on the three output channels in Table~\ref{tab:nominal_performance}. For brevity, we exclusively report the results obtained on the test dataset denoted as ``FullMSine\_Level6\_Validation''.
\begin{table}
\centering
\captionsetup{justification=centering} % Centering captions
%\small % Reduced font size
\setlength{\tabcolsep}{4pt} % Reduced column padding
\adjustbox{scale=0.7}{
\begin{tabular}{@{}l*{9}{c}@{}} % Adjusting column alignment
\toprule
\multirow{2}{*}{Regularization} & \multicolumn{3}{c}{Channel 1} & \multicolumn{3}{c}{Channel 2} & \multicolumn{3}{c}{Channel 3} \\
\cmidrule(lr){2-4} \cmidrule(lr){5-7} \cmidrule(lr){8-10}
& {\emph{fit}} & {RMSE} & {NRMSE} & {\emph{fit}} & {RMSE} & {NRMSE} & {\emph{fit}} & {RMSE} & {NRMSE} \\
\midrule
No reg. & 86.5 & 0.180 & 0.134 & 90.0 & 0.167 & 0.099  & 76.2 & 0.368 & 0.237 \\
Modal $\ell_1$ & 85.4 & 0.195 & 0.145 & 89.8 & 0.171 & 0.102 & 74.5 & 0.395 & 0.254 \\
Hankel norm & 85.8 & 0.190 & 0.142 & 89.0 & 0.185 & 0.110 & 75.5 & 0.379 & 0.245 \\
\bottomrule
\end{tabular}
}
\caption{Performance of the SSM trained with different regularization methods.}
\label{tab:nominal_performance}
\end{table}
The three trained models achieve similar performance, which is also in line with existing state-of-the-art.
For instance, the average NRMSE over the three channels is about 0.15, which is close to the
result reported in~\cite{revay2023recurrent}.
%\footnote{The results in \cite{revay2023recurrent} are reported in a figure and cannot be determined with great precision.}
However, we observe that regularization has a strong effect on the properties of the estimated LTI blocks.

\begin{figure*}[ht]
    \centering
    %\includegraphics[scale=.99]{figures/F16/ckpt_large_no_reg_eigs_complex.pdf}
    %\vskip -1em
    \includegraphics[scale=.70]{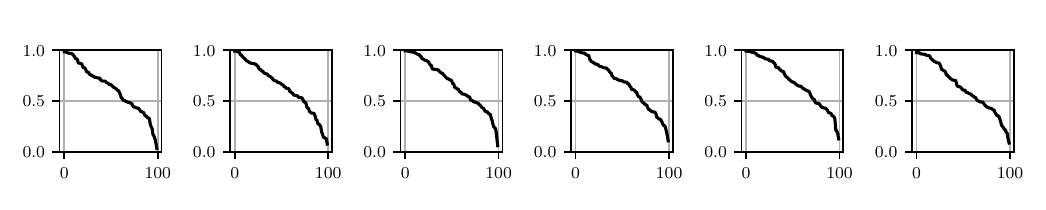}
    \vskip -1em
    \includegraphics[scale=.70]{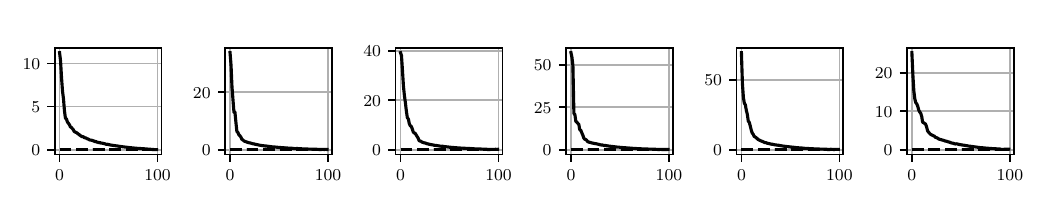}
    % \caption{No regularization: complex eigenvalues in the unit circle (top), eigenvalues magnitude (middle) and Hankel singular values (bottom).}
    \caption{No regularization: eigenvalues magnitude (top) and Hankel singular values (bottom).}
    \label{fig:lru_no_reg}
\end{figure*}

\begin{figure*}[h]
    \centering
    %\includegraphics[scale=.99]{figures/F16/ckpt_large_reg_modal_eigs_complex.pdf}
    %\vskip -1em
    \includegraphics[scale=.70]{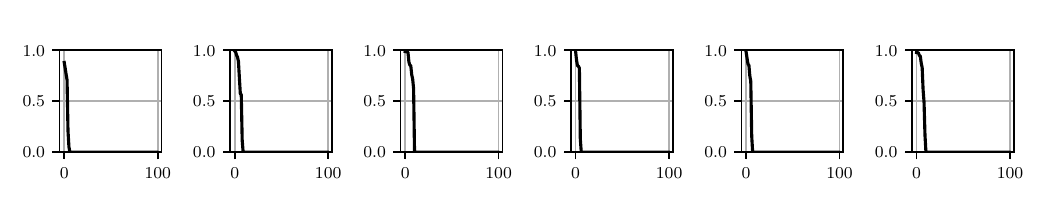}
    \vskip -1em
    \includegraphics[scale=.70]{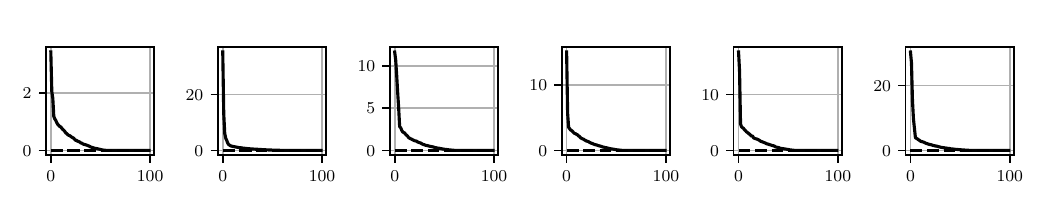}
    % \caption{Modal LASSO regularization: complex eigenvalues in the unit circle (top), eigenvalues magnitude (middle), and hankel singular values (bottom).}
    \caption{Modal $\ell_1$ regularization: eigenvalues magnitude (top) and Hankel singular values (bottom).}
    \label{fig:lru_modal_reg}
\end{figure*}

\begin{figure*}[h]
    \centering
    %\includegraphics[scale=.99]{figures/F16/ckpt_large_reg_hankel_eigs_complex.pdf}
    %\vskip -1em
    \includegraphics[scale=.70]{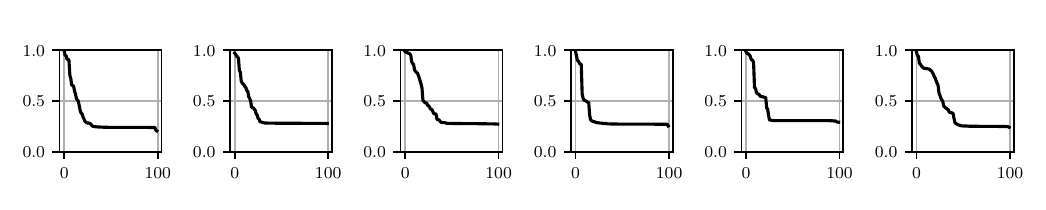}
    \vskip -1em
    \includegraphics[scale=.70]{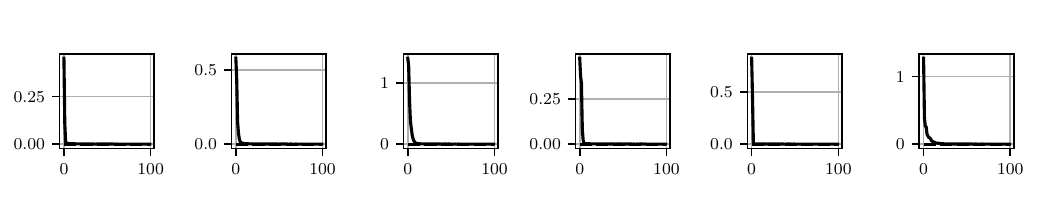}
    %\caption{Hankel Nuclear-norm regularization: complex eigenvalues in the unit circle (top), eigenvalues magnitude (middle), and hankel singular values (bottom).}
    \caption{Hankel nuclear norm regularization: eigenvalues magnitude (top) and Hankel singular values (bottom).}
    \label{fig:lru_hankel_reg}
\end{figure*}

The plots in the six columns of Fig.~\ref{fig:lru_no_reg}, \ref{fig:lru_modal_reg} and \ref{fig:lru_hankel_reg}  illustrate this effect, where each column corresponds to LRU in one of the $6$ layers.  
For the model without regularization (Fig.~\ref{fig:lru_no_reg}), most of the eigenvalues have non-zero magnitude (top panel). In this sense, the {modal} reduction methods MT/MSP are not expected to be effective. The Hankel singular values decrease  slightly faster towards zero, suggesting that the effectiveness of the balanced reduction methods BT/BSP might be marginally superior.
As for the model obtained with modal $\ell_1$-regularization (Fig.~\ref{fig:lru_modal_reg}), several eigenvalues have been  pushed towards zero (top panel), suggesting the potential effectiveness of modal reduction methods. 
%Conversely, the Hankel singular values are only marginally faster decreasing towards zero (bottom panel).
Finally, for the model trained with Hankel nucelar norm regularization (Fig.~\ref{fig:lru_hankel_reg}), the Hankel singular values decrease very sharply towards zero (bottom panel), while none of the eigenvalues' magnitude is pushed towards zero. Thus, we expect balanced reduction methods to be effective in this case.

\begin{figure}
\centering
\includegraphics[scale=.8]{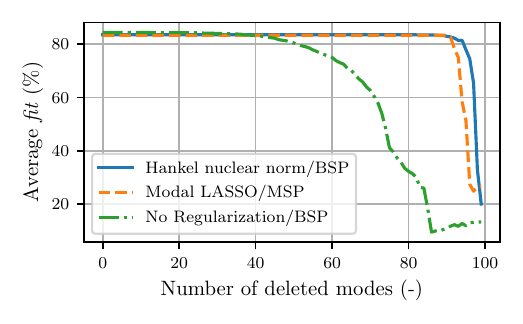}
\caption{Test performance obtained for increasing number of removed modes in all layers, for selected combinations of regularization and model order reduction approaches.}
\label{fig:fit_vs_removed}
\end{figure}

In Table \ref{tab:regularization}, we report the maximum number of eliminated states  with the different MOR techniques applied to the three  trained models, such that the performance degradation in test (in terms of \emph{fit} index) averaged over the three output channels is less than 1\%. The best results are obtained for the combinations of modal $\ell_1$-regularization followed by MSP and Hankel nuclear norm regularization followed by BSP, which both lead to 91 eliminated states. We also observe that, when regularization is not applied in training, the subsequent MOR is decisively less effective. Fig.~\ref{fig:fit_vs_removed} further highlights this key experimental results: when training with no regularization, the best reduction approach (BSP) is significantly less effective than the optimal regularizer+MOR combinations: modal $\ell_1$+MSP and Hankel nuclear norm+BSP.

\begin{table}%[htbp]
    \centering
    \begin{tabular}{@{}lllll@{}}
    \toprule
    & \multicolumn{4}{c}{Truncation Method} \\ \cmidrule(lr){2-5}
    Regularization Method & BT & BSP & MT & MSP \\ \midrule
    No Regularization & 28 & 43 & 3 & 35 \\
    Modal $\ell_1$& 56 & 73 & 0 & \textbf{91} \\
    Hankel nuclear norm& 89 & \textbf{91} & 18 & 76 \\ \bottomrule
    \end{tabular}
    \caption{Maximum number of modes that can be eliminated while keeping the performance of the trained model within 1\% of the full case.}
    \label{tab:regularization}
\end{table}

\FloatBarrier

\section{Conclusions}
We have presented regularization methods and model order reduction approaches that enable substantial
 simplification of deep structured state-space models. Our experimental results suggest that regularization is a fundamental ingredient of our procedure. Indeed, model order reduction executed  as a mere post-hoc step, after a standard training conducted without regularization appears to be significantly less effective. In future works, we will analyze in more depth the effect of the regularization strength $\gamma$ through more extensive numerical experiments and possibly with analytical tools. Moreover, we aim at decreasing the number of internal input/output channels $d_{\rm model}$ and of layers $n_{\rm layers}$ of the architecture. A possible approach is based on group LASSO regularization, following an approach similar to the one recently introduced in~\cite{bemporad2024linear} for more classical neural state-space models. Finally, we will extend our techniques to other architectures that feature linear dynamical blocks at their core, such as dynoNet and deep Koopman representations.
\appendix
\section{Appendix}
\subsection{Computation of Hankel singular values}
\label{sec:app}
The Hankel singular values of a discrete-time LTI system with complex-valued matrices $(A, B, C, D)$ are given by:
\begin{equation}
\sigma_j = \sqrt{\eig_{j}(PQ)}, \ \ \forall j \in \range{1}{\nx},
\end{equation}
where $P \in \mathbb{C}^{\nx \times \nx}$ and $Q \in \mathbb{C}^{\nx \times \nx}$ are the controllability and observability 
Grammians, respectively, which are the solutions of the discrete Lyapunov equations~\cite{katayama05}:
\begin{align} 
APA^{*} - P + BB^{*} &= 0\\
A^{*}QA - Q + C^{*}C &= 0,
\end{align}
%where the ${}^*$ denotes the conjugate transpose (Hermetian) of a matrix.

\subsection{Solution to a diagonal Discrete Lyapunov equation}
We show that discrete Lyapunov equations can be solved  efficiently for systems in modal representation where matrix $A$ is diagonal.
The direct method to solve the Lyapunov equation with variable $X$:
\begin{equation}
    \label{eq:dlyap_generic}
    AXA^{*} - X + Y = 0
\end{equation}
is obtained by exploiting the product property:
\begin{equation}
\vec(A X B) = (B^\top \otimes A) \vec(X),
\end{equation}
where $\otimes$ is the Kronecker product operator and $\vec(\cdot)$ represents the \emph{column-wise} vectorization operation.
Applying this formula to \eqref{eq:dlyap_generic}, one obtains:
\begin{equation}
\label{eq:direct}
(I - A^* \otimes A) \vec (X) = \vec (Y),
\end{equation}
which is a linear system in the unknowns $\vec (X)$. If $A$ is diagonal, the left-hand side matrix of \eqref{eq:direct} is also diagonal, and thus its solution is simply obtained through $n_x^2$ scalar divisions.

%\section*{Acknowledgements}

\bibliographystyle{plain}
\bibliography{ms}

\end{document}